\documentclass[11pt,a4paper]{article}
\usepackage[hyperref]{emnlp2018}
\usepackage{times}
\usepackage{latexsym}

\usepackage{url}

\usepackage{amsmath}
\usepackage{amssymb}
\usepackage{amsopn}
\usepackage{bbding}
\usepackage{subcaption}
\usepackage{CJKutf8}
\usepackage{graphicx}
\usepackage{multirow}
\usepackage{color}
\usepackage{tabularx}
\usepackage{enumitem}
\usepackage{algorithm}  
\usepackage{tikz}
\usepackage{algorithmic}  
\usepackage{caption}

\usepackage{verbatim}

\usepackage{enumitem}

\aclfinalcopy % Uncomment this line for the final submission

\title{Simplifying Neural Machine Translation with Addition-Subtraction Twin-Gated Recurrent Networks}

\author{Biao Zhang$^{1}$, Deyi Xiong$^{2}$, Jinsong Su$^{1}$\thanks{Corresponding author.}, Qian Lin$^{1}$  \textmd{and} Huiji Zhang$^{3}$\\
	Xiamen University, Xiamen, China$^{1}$ \\
	Soochow University, Suzhou, China$^{2}$ \\
	Xiamen Meiya Pico information Co.,Ltd. Xiamen, China$^{3}$ \\
	{\tt \{zb,qianl\}@stu.xmu.edu.cn, dyxiong@suda.edu.cn, jssu@xmu.edu.cn} \\
	{\tt  zhanghj@300188.cn}
}

\date{}

\begin{document}
\maketitle
\begin{abstract}
In this paper, we propose an addition-subtraction twin-gated recurrent network (ATR) to simplify neural machine translation. The recurrent units of ATR are heavily simplified to have the smallest number of weight matrices among units of all existing gated RNNs. With the simple addition and subtraction operation, we introduce a twin-gated mechanism to build input and forget gates which are highly correlated. Despite this simplification, the essential non-linearities and capability of modeling long-distance dependencies are preserved. Additionally, the proposed ATR is more transparent than LSTM/GRU due to the simplification. Forward self-attention can be easily established in ATR, which makes the proposed network interpretable. 
Experiments on WMT14 translation tasks demonstrate that ATR-based neural machine translation can yield competitive performance on English-German and English-French language pairs in terms of both translation quality and speed. Further experiments on NIST Chinese-English translation, natural language inference and Chinese word segmentation verify the generality and applicability of ATR on different natural language processing tasks.\footnote{Source code is available at: https://github.com/bzhangGo/ATR and https://github.com/bzhangGo/zero}
\end{abstract}

\section{Introduction}
Neural machine translation (NMT), typically with an attention-based encoder-decoder framework~\citep{DBLP:journals/corr/BahdanauCB14}, has recently become the dominant approach to machine translation and already been deployed for online translation services~\cite{DBLP:journals/corr/WuSCLNMKCGMKSJL16}. Recurrent neural networks (RNN), e.g., LSTMs~\citep{Hochreiter:1997:LSM:1246443.1246450} or GRUs~\citep{journals/corr/ChungGCB14}, are widely used as the encoder and decoder for NMT. In order to alleviate the gradient vanishing
issue found in simple recurrent neural networks (SRNN)~\cite{elman1990finding}, recurrent units in LSTMs or GRUs normally introduce different gates to create shotcuts for gradient information to pass through. 

Notwithstanding the capability of these gated recurrent networks in learning long-distance dependencies, they use remarkably more matrix transformations (i.e., more parameters) than SRNN. And with many non-linear functions modeling inputs, hidden states and outputs, they are also less transparent than SRNN. These make NMT which is based on these gated RNNs suffer from not only inefficiency in training and inference due to recurrency and heavy computation in recurrent units~\cite{DBLP:journals/corr/VaswaniSPUJGKP17} but also difficulty in producing interpretable models~\cite{DBLP:journals/corr/LeeLZ17}. These also hinder the deployment of NMT models particularly on memory- and computation-limited devices.

In this paper, our key interest is to simplify recurrent units in RNN-based NMT. In doing so, we want to investigate how further we can advance RNN-based NMT in terms of the number of parameters (i.e., memory consumption), running speed and interpretability. This simplification shall preserve the capability of modeling long-distance dependencies in LSTMs/GRUs and the expressive power of recurrent non-linearities in SRNN. The simplification shall also reduce computation load and physical memory consumption in recurrent units on the one hand and allow us to take a good look into the inner workings of RNNs on the other hand. 

In order to achieve this goal, we propose an addition-subtraction twin-gated recurrent network (ATR) for NMT. In the recurrent units of ATR, we only keep the very essential weight matrices: one over the input and the other over the history (similar to SRNN). Comparing with previous RNN variants (e.g., LSTM or GRU), we have the smallest number of weight matrices. This will reduce the computation load of matrix multiplication. ATR also uses gates to bypass the vanishing gradient problem so as to capture long-range dependencies. Specifically, we use the addition and subtraction operations between the weighted history and input to estimate an  input and forget gate respectively. These add-sub operations not only distinguish the two gates so that we do not need to have different weight matrices for them, but also make the two gates dynamically correlate to each other. Finally, we remove some non-linearities in recurrent units. 

Due to these simplifications, we can easily show that each new state in ATR is an unnormalized weighted sum of previous inputs, similar to recurrent additive networks~\cite{DBLP:journals/corr/LeeLZ17}. This property not only allows us to trace each state back to those inputs which contribute more but also establishes unnormalized forward self-attention between the current state and all its previous inputs. The self-attention mechanism has already proved very useful in non-recurrent NMT~\cite{DBLP:journals/corr/VaswaniSPUJGKP17}.

We build our NMT systems on the proposed ATR with a single-layer encoder and decoder. Experiments
on WMT14 English-German and English-French
translation tasks show that our model yields competitive results compared with GRU/LSTM-based NMT. When we integrate an orthogonal context-aware encoder (still single layer) into ATR-based NMT, our model (yielding 24.97 and 39.06 BLEU on English-German and English-French translation respectively) is even comparable to deep RNN and non-RNN NMT models which are all with multiple encoder/decoder layers. In-depth analyses demonstrate that ATR is more efficient than LSTM/GRU in terms of NMT training and decoding speed. 

We adapt our model to other language translation and natural language processing tasks, including NIST Chinese-English translation, natural language inference and Chinese word segmentation. Our conclusions still hold on all these tasks.

\section{Related Work}

The most widely used RNN models are LSTM~\citep{Hochreiter:1997:LSM:1246443.1246450} and GRU~\citep{journals/corr/ChungGCB14}, both of which are good at handling gradient vanishing problem, a notorious bottleneck of the simple RNN~\cite{elman1990finding}. The design of gates in our model follows the gate philosophy in LSTM/GRU. 

Our work is closely related to the recurrent additive network (RAN) proposed by \citet{DBLP:journals/corr/LeeLZ17}. They empirically demonstrate that many non-linearities commonly used in RNN transition dynamics can be removed, and that recurrent hidden states computed as purely the weighted sum of input vectors can be quite efficient in language modeling. Our work follows the same spirit  of simplifying recurrent units as they do. But our proposed ATR is significantly different from RAN in three aspects. First, ATR is simpler than RAN with even fewer parameters. There are only two weight matrices in ATR while four different weight matrices in the simplest version of RAN (two for each gate in RAN). Second, since the only difference between the input and forget gate in ATR is the addition/subtraction operation between the history and input, the two gates can be learned to be highly correlated as shown in our analysis. Finally, although RAN is verified effective in language modeling, our experiments show that ATR is better than RAN in machine translation in terms of both speed and translation quality.

\begin{figure*}[t]
\centering
\captionsetup[subfigure]{skip=12pt,belowskip=3pt,aboveskip=6pt}
\begin{subfigure}{0.3\textwidth}
\includegraphics[scale=0.52]{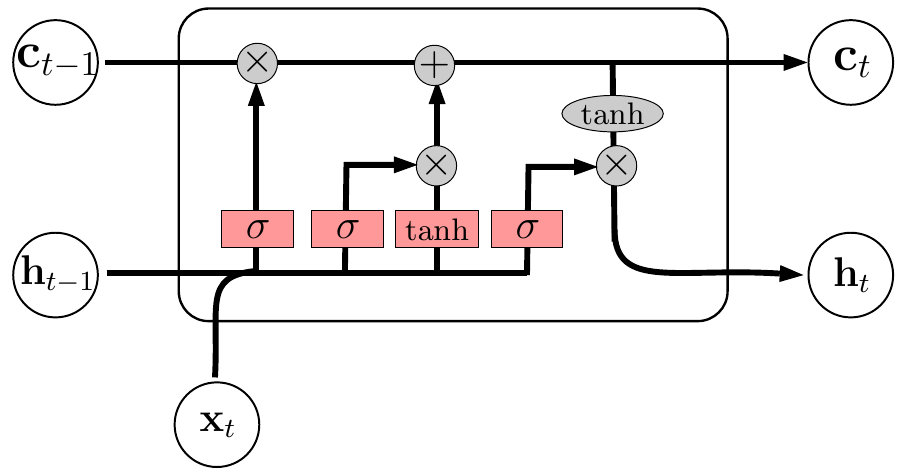}
\caption{\label{lstm} LSTM}
\end{subfigure}~
\begin{subfigure}{0.3\textwidth}
\includegraphics[scale=0.52]{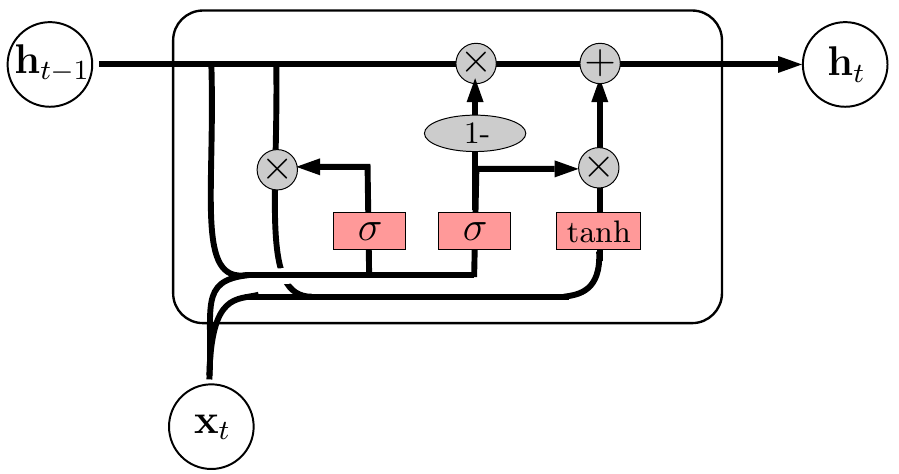}
\caption{\label{gru} GRU}
\end{subfigure}~
\begin{subfigure}{0.3\textwidth}
\includegraphics[scale=0.52]{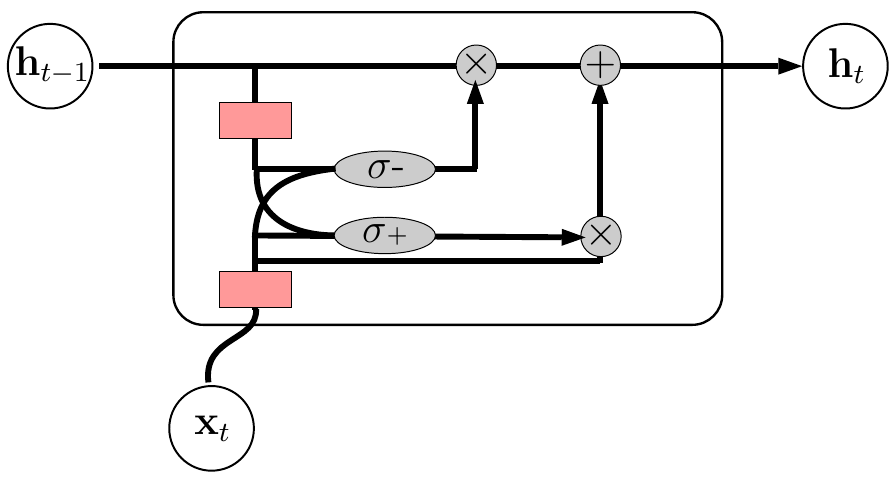}
\caption{\label{eru} ATR}
\end{subfigure}
\caption{\label{models} Architecture for LSTM, GRU and ATR. $\mathbf{c}_{*}$ indicates the memory cell specific to the LSTM network. $\mathbf{x}_{*}$ and $\mathbf{h}_{*}$ denote the input and output hidden states respectively.}
\end{figure*}
To speed up RNN models, a line of work has attempted to remove recurrent connections. For example, \citet{DBLP:journals/corr/BradburyMXS16} propose the quasi-recurrent neural network (QRNN) which uses convolutional layers and a minimalist recurrent pooling function to improve parallelism. 
Very recently, \citet{2017arXiv170902755L} propose a simple recurrent unit (SRU). With the cuDNN optimization, their RNN model can be trained as fast as CNNs. However, to obtain promising results, QRNN and SRU have
to use deep architectures. In practice, 4-layer
QRNN encoder and decoder are used to gain
translation quality that is comparable to that of
single-layer LSTM/GRU NMT.
In particular, our one-layer model achieves significantly higher performance than a 10-layer SRU system.

Finally, our work is also related to the efforts in developing alternative architectures for NMT models. \citet{DBLP:journals/corr/ZhouCWLX16} introduce fast-forward connections between adjacent stacked RNN layers to ease gradient propagation. \citet{wang-EtAl:2017:Long1} propose a linear associate unit to reduce the gradient propagation length along layers in deep NMT. \citet{DBLP:journals/corr/GehringAGYD17} and \citet{DBLP:journals/corr/VaswaniSPUJGKP17} explore purely convolutional and attentional architectures as alternatives to RNNs for neural translation. With careful configurations, their deep models achieve state-of-the-art performance on various datasets.

\section{Addition-Subtraction Twin-Gated Recurrent Network}

Given a sequence $\mathbf{x}=\{\mathbf{x}_1, \mathbf{x}_2,$$ \ldots, \mathbf{x}_T\}$, RNN updates the hidden state $\mathbf{h}_t$ recurrently as follows:
\begin{equation}
\mathbf{h}_t = \phi (\mathbf{h}_{t-1}, \mathbf{x}_t)
\end{equation}
where $\mathbf{h}_{t-1}$ is the previous hidden state, which is considered to store information from all previous inputs, and $\mathbf{x}_t$ is the current input. The function $\phi(\cdot)$ is a non-linear recurrent function, abstracting away from details in recurrent units. 

GRU can be considered as a simplified version of LSTM. In this paper, theoretically, we use GRU as our benchmark and propose a new recurrent unit to further simplify it. The GRU function is defined as follows (see Figure \ref{gru}):
\begin{align}%\label{eq_gru}
\mathbf{z}_t = \sigma(\mathbf{W}_z \mathbf{x}_t + \mathbf{U}_z \mathbf{h}_{t-1}) \label{gru_z} \\
\mathbf{r}_t = \sigma(\mathbf{W}_r \mathbf{x}_t + \mathbf{U}_r \mathbf{h}_{t-1}) \label{gru_r} \\
\tilde{\mathbf{h}}_t = \tanh(\mathbf{W}_h \mathbf{x}_t + \mathbf{U}_h (\mathbf{r}_t \odot \mathbf{h}_{t-1}))\label{gru_h_} \\
\mathbf{h}_t = \mathbf{z}_t \odot \mathbf{h}_{t-1} + (1 - \mathbf{z}_t) \odot \tilde{\mathbf{h}}_t \label{gru_h}
\end{align}
where $\odot$ denotes an element-wise multiplication. The {\em reset} gate $\mathbf{r}_t$ and {\em update} gate $\mathbf{z}_t$ enable manageable information flow from the history and the current input to the new state respectively. Despite the success of these two gates in handling gradient flow, they consume extensive matrix transformations and weight parameters.

We argue that many of these matrix transformations are not essential. We therefore propose an addition-subtraction twin-gated recurrent unit (ATR), formulated as follows (see Figure \ref{eru}):
\begin{align}\label{eq_eru}
\mathbf{q}_t = \mathbf{W}_h \mathbf{h}_{t-1},~\mathbf{p}_t = \mathbf{W}_x \mathbf{x}_t \\  
\mathbf{i}_t = \sigma(\mathbf{p}_t + \mathbf{q}_t) \label{twin_i}\\
\mathbf{f}_t = \sigma(\mathbf{p}_t - \mathbf{q}_t) \label{twin_f}\\
\mathbf{h}_t = \mathbf{i}_t \odot \mathbf{p}_t + \mathbf{f}_t \odot \mathbf{h}_{t-1} \label{twin_h}
\end{align}
The hidden state $\mathbf{h}_t$ in ATR is a weighted mixture of both the current input $\mathbf{p}_t$ and the history $\mathbf{h}_{t-1}$ controlled by an {\em input} gate $\mathbf{i}_t$ and a {\em forget} gate $\mathbf{f}_t$ respectively. Notice that we use the transformed representation $\mathbf{p}_t$ for the current input rather than the raw vector $\mathbf{x}_t$ due to the potential mismatch in dimensions between $\mathbf{h}_t$ and $\mathbf{x}_t$. 

Similar to GRU, we use gates, especially the forget gate, to control the back-propagated gradient flow to make sure gradients will neither vanish nor explode. We also preserve the non-linearities of SRNN in ATR but only in the two gates. 

There are three significant differences of ATR from GRU. Some of these differences are due to the simplifications introduced in ATR. First, we squeeze the number of weight matrices in gate calculation from four to two (see Equation (\ref{gru_z}\&\ref{gru_r}) and (\ref{twin_i}\&\ref{twin_f})). In all existing gated RNNs, the inputs to gates are weighted sum of the previous hidden state and input. In order to distinguish these gates, the weight matrices over the previous hidden state and the current input should be different for different gates. The number of different weight matrices in gates is therefore 2$|$\#gates$|$ in previous gated RNNs. Different from them, ATR introduces different operations (i.e., addition and subtraction) between the weighted history and input to distinguish the input and forget gate. Therefore, the weight matrices over the previous state/input in the two gates can be the same in ATR. Second, we keep the very essential non-linearities, only in the two gates. In ATR, the role of $\mathbf{p}_t$ is similar to that of $\tilde{\mathbf{h}}_t$ in GRU (see Equation (\ref{gru_h_})). However, we completely wipe out the recurrent non-linearity of $\tilde{\mathbf{h}}_t$ in $\mathbf{p}_t$ (i.e., $\mathbf{p}_t = \mathbf{W}_x \mathbf{x}_t$). \newcite{DBLP:journals/corr/LeeLZ17} show that this non-linearity is not necessary in language modeling. We further empirically demonstrate that it is neither essential in machine translation. Third, in GRU the gates for $\tilde{\mathbf{h}}_t$ and $\mathbf{h}_{t-1}$ are coupled and normalized to 1 while we do not explicitly associate the two gates in ATR. Instead, they can be learned to be correlated in an implicit way, as shown in the next subsection and our empirical analyis in Section \ref{exp_analysis_twin}.

\subsection{Twin-Gated Mechanism}\label{twin_mechanism}

Unlike GRU, we use an addition and subtraction operation over the transformed current input $\mathbf{p}_t$ and history $\mathbf{q}_t$ to differentiate the two gates in ATR. As the two gates have the same weights for their input components with only a single difference in the operation between the input components, they act like twins. We term the two gates in ATR as twin gates and the procedure, shown in Equation (\ref{twin_i}\&\ref{twin_f}), as the {\em twin-gated mechanism}. This mechanism endows our model with the following two advantages: 1) Both addition and subtraction operations are completely linear so that fast computation can be expected; and 2) No other weight parameters are introduced for gates so that our model is more memory-compact.

\begin{figure}[t]
\centering
\includegraphics[scale=0.72]{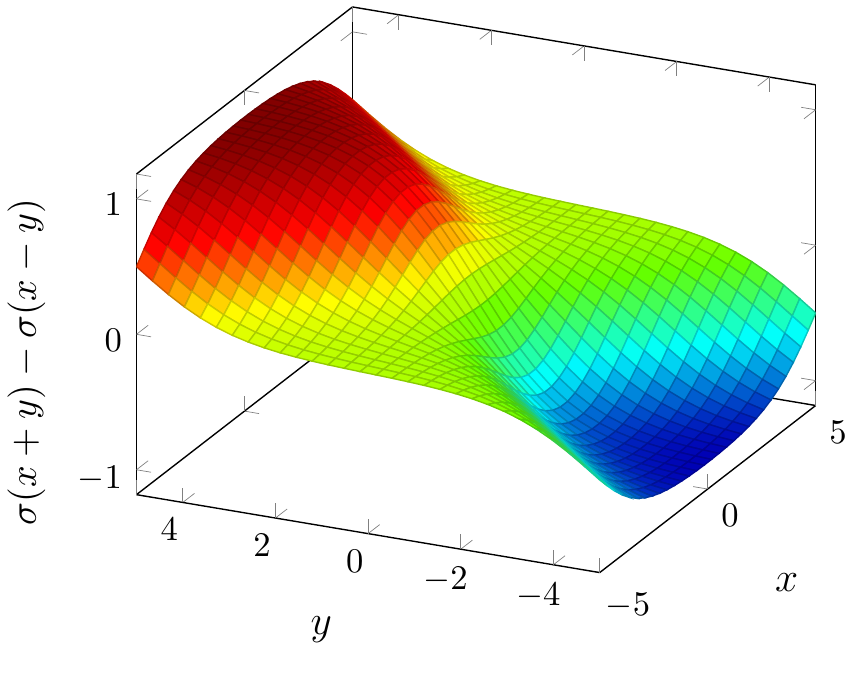}
\caption{\label{twin_gate_vis} Visualization of the difference between $\sigma(x+y)$ (the input gate) and $\sigma(x-y)$ (the forget gate). Here, we set $x,y \in [-5, 5]$.}
\end{figure}

A practical question for the twin-gated mechanism is whether twin gates are really capable of dynamically weighting the input and history information. To this end, we plot the surface of one-dimensional $\sigma(x+y)-\sigma(x-y)$ in Figure \ref{twin_gate_vis}. It is clear that both gates are highly non-linearly correlated, and that there are regions where $\sigma(x+y)$ is equal to, greater or smaller than $\sigma(x-y)$. In other words, by adapting the distribution of input and forget gates, the twin-gated mechanism has the potential to automatically seek suitable regions in Figure \ref{twin_gate_vis} to control its preference between the new and past information. We argue that the input and forget gates are negatively correlated after training, and empirically show their actual correlation in Section \ref{exp_analysis_twin}.

\subsection{Computation Analysis}\label{com_analysis}

Here we provide a systematical comparison of computations in LSTM, GRU, RAN and our ATR with respect to the number of weight matrices and matrix transformations. Notice that all these units are building blocks of RNNs so that the total computational complexity and the minimum number of sequential operations required are unchanged, i.e. $\mathcal{O}(n\cdot d^2)$ and $\mathcal{O}(n)$ respectively where $n$ is the sequence length and $d$ is the dimensionality of hidden states. However, the actual number of matrix transformations in the unit indeed significantly affects the running speed of RNN in practice.

We summarize the results in Table \ref{computation_analysis}. LSTM contains three different gates and a cell state, including 4 different neural layers with 8 weight matrices and transformations. GRU simplifies LSTM by removing a gate, but still involves two gates and a candidate hidden state. It includes 3 different neural layers with 6 weight matrices and transformations. RAN further simplifies GRU by removing the non-linearity in the state transition and therefore contains 4 weight matrices in its simplest version. Although our ATR also has two gates, however, there are only 2 weight matrices and transformations, accounting for only a third and a quarter of those in GRU and LSTM respectively. To the best of our knowledge, ATR has the smallest number of weight transformations in existing gated RNN units. We provide a detailed and empirical analysis on the speed in Section \ref{exp_analysis_speed}.

\begin{table}[t]
\begin{center}
{\small
\begin{tabular}{c|c|c}
{\bf Model} & {\bf \# WM} & {\bf \# MT} \\
\hline
\hline
{\it LSTM} & 8 & 8 \\
{\it GRU} & 6 & 6 \\
{\it RAN} & 4 & 4 \\
{\it ATR} & 2 & 2 \\
\end{tabular}
}
\end{center}
\caption{\label{computation_analysis} Comparison of LSTM, GRU, RAN and ATR in terms of the number of weight matrices (WM) and matrix transformations (MT).}
\end{table}
\subsection{Interpretability Analysis of Hidden States}\label{dependency_analysis}

An appealing property of the proposed ATR is its interpretability. This can be demonstrated by rolling out Equation (\ref{twin_h}) as follows:
\begin{equation}\label{dep_gate_value}
\begin{split}
\mathbf{h}_t & = \mathbf{i}_t \odot \mathbf{p}_t + \mathbf{f}_t \odot \mathbf{h}_{t-1} \\
			    & = \mathbf{i}_t \odot \mathbf{W}_x \mathbf{x}_t + \sum_{k=1}^{t-1} \mathbf{i}_k \odot \left(\prod_{l=1}^{t-k}
			     \mathbf{f}_{k+l}\right) \odot \mathbf{W}_x \mathbf{x}_k \\
				& \approx \sum_{k=1}^t \mathbf{g}_k \odot \mathbf{W}_x \mathbf{x}_k
\end{split}
\end{equation}
where $\mathbf{g}_k$ can be considered as an approximate weight assigned to the $k$-th input. Similar to the RAN model~\cite{DBLP:journals/corr/LeeLZ17}, the hidden state in ATR is a component-wise weighted sum of the inputs. This not only enables ATR to build up essential dependencies between preceding inputs and the current hidden state, but also allows us to easily detect which previous words have the promising impacts on the current state. This desirable property obviously makes ATR highly interpretable. 

Additionally, this form of weighted sum is also related to self-attention~\cite{DBLP:journals/corr/VaswaniSPUJGKP17}. It can be considered as a forward unnormalized self-attention where each hidden state attends to all its previous positions. As the self-attention mechanism has proved very useful in NMT~\cite{DBLP:journals/corr/VaswaniSPUJGKP17}, we conjecture that such property of ATR partially contributes to its success in machine translation as shown in our experiments. We visualize the dependencies captured by Equation (\ref{dep_gate_value}) in Section \ref{exp_analysis_dep}.

\begin{table*}[t]
\begin{center}
{ \small
\begin{tabular}{l|l|c|c|c}
\multicolumn{1}{l|}{\bf System} &
\multicolumn{1}{l|}{\bf Architecture } &
\multicolumn{1}{c|}{\bf Vocab} &
\multicolumn{1}{c|}{\bf {\em tok} BLEU } & 
\multicolumn{1}{c}{\bf {\em detok} BLEU}\\
\hline
\hline
\citet{BUCK14.1097.L14-1074} & WMT14 winner system phrase-based + large LM  & - & - & 20.70 \\
\hline
\multicolumn{5}{c}{\it Existing deep NMT systems (perhaps different tokenization)} \\
\hline
\citet{DBLP:journals/corr/ZhouCWLX16} & LSTM with 16 layers + F-F connections & 160K & 20.60 & - \\
\citet{2017arXiv170902755L} & SRU with 10 layers & 50K & 20.70 & - \\
\citet{2018arXiv180504185A} & SR-NMT with 4 layers & 32K & 23.32 & - \\
\citet{wang-EtAl:2017:Long1} & GRU with 4 layers + LAU + PosUnk & 80K & 23.80 & - \\
\citet{wang-EtAl:2017:Long1} & GRU with 4 layers + LAU + PosUnk + ensemble & 80K & 26.10 & - \\
\citet{DBLP:journals/corr/WuSCLNMKCGMKSJL16} & LSTM with 8 layers + RL-refined WPM & 32K & 24.61 & - \\
\citet{DBLP:journals/corr/WuSCLNMKCGMKSJL16} & LSTM with 8 layers + RL-refined ensemble & 80K & 26.30 & - \\
\citet{DBLP:journals/corr/VaswaniSPUJGKP17} & Transformer with 6 layers + base model & 37K & 27.30 & - \\
\hline
\multicolumn{5}{c}{\it Comparable NMT systems (the same tokenization)} \\
\hline
\citet{luong-pham-manning:2015:EMNLP} & LSTM with 4 layers + local att. + unk replace & 50K & 20.90 & - \\
\citet{DBLP:journals/corr/ZhangXS17} & GRU with gated attention + BPE & 40K & 23.84 & - \\
\citet{DBLP:journals/corr/GehringAGYD17} & CNN with 15 layers + BPE & 40K & 25.16 & - \\
\citet{DBLP:journals/corr/GehringAGYD17} & CNN with 15 layers + BPE + ensemble & 40K & 26.43 & - \\
\citet{P18-1166} & Transformer with 6 layers + aan + base model & 32K & 26.31 & - \\
\hline
\multicolumn{5}{c}{\it Our end-to-end NMT systems} \\
\hline
\multirow{7}{*}{\it this work} & RNNSearch + GRU + BPE & 40K & 22.54 & 22.06 \\
									  & RNNSearch + LSTM + BPE & 40K & 22.96 & 22.39 \\
									  & RNNSearch + RAN + BPE & 40K & 22.14 & 21.40  \\
									  & RNNSearch + ATR + BPE & 40K & 22.48 & 21.99 \\
									  & RNNSearch + ATR + CA + BPE & 40K & 23.31 & 22.70 \\
									  & GNMT + ATR + BPE & 40K & 24.16 & 23.59 \\
									  & RNNSearch + ATR + CA + BPE + ensemble & 40K & 24.97 & 24.33 \\
\end{tabular}
}
\end{center}
\caption{\label{english_german_translation} Tokenized ({\em tok}) and detokenized (detok) case-sensitive BLEU scores on the WMT14 English-German translation task. ``{\it unk replace}'' and ``{\it PosUnk}'' denotes the approach of handling rare words in~\citet{jean-EtAl:2015:ACL-IJCNLP} and~\citet{luong-pham-manning:2015:EMNLP} respectively. ``{\em RL}'' and ``{\em WPM}'' is the reinforcement learning optimization and word piece model used in~\citet{DBLP:journals/corr/WuSCLNMKCGMKSJL16}. ``{\em CA}'' is the context-aware recurrent encoder~\cite{8031316}. ``{\em LAU}'' and ``{\em F-F}'' denote the linear associative unit and the fast-forward architecture proposed by~\citet{wang-EtAl:2017:Long1} and~\citet{DBLP:journals/corr/ZhouCWLX16} respectively. ``{\em aan}'' denotes the average attention network proposed by ~\citet{P18-1166}.
}
\end{table*}
\section{Experiments}

\subsection{Setup}
We conducted our main experiments on WMT14 English-German and English-French translation tasks. Translation quality is measured by case-sensitive BLEU-4 metric~\citep{PapineniEtAl2002}. Details about each dataset are as follows:
\begin{description}
\item[English-German] To compare with previous reported results~\citep{luong-EtAl:2015:ACL-IJCNLP,jean-EtAl:2015:ACL-IJCNLP,DBLP:journals/corr/ZhouCWLX16,wang-EtAl:2017:Long1}, we used the same training data of WMT 2014, which consist of 4.5M sentence pairs. We used the newstest2013 as our dev set, and the newstest2014 as our test set.
\item[English-French] We used the WMT 2014 training data. This corpora contain 12M selected sentence pairs. We used the concatenation of newstest2012 and newstest2013 as our dev set, and the newstest2014 as our test set.
\end{description}

The used NMT system is an attention-based encoder-decoder system, which employs a bidirectional recurrent network as its encoder and a two-layer hierarchical unidirectional recurrent network as its decoder, companied with an additive attention mechanism~\cite{DBLP:journals/corr/BahdanauCB14}. We replaced the recurrent unit with our proposed ATR model. More details are given in Appendix~\ref{nmt_atr}. 

We also conducted experiments on Chinese-English translation, natural language inference and Chinese word segmentation. Details and experiment results are provided in Appendix~\ref{other_exp}.

\subsection{Training}

We set the maximum length of training instances to 80 words for both English-German and English-French task. We used the {\em byte pair encoding} compression algorithm~\citep{sennrich-haddow-birch:2016:P16-12} to reduce the vocabulary size as well as to deal with the issue of rich morphology. We set the vocabulary size of both source and target languages to 40K for all translation tasks. All out-of-vocabulary words were replaced with a token ``{\it unk}''.

We used 1000 hidden units for both encoder and decoder. All word embeddings had dimensionality 620. We initialized all model parameters randomly according to a uniform distribution ranging from -0.08 to 0.08. These tunable parameters were then optimized using Adam algorithm~\citep{DBLP:journals/corr/KingmaB14} with the two momentum parameters set to 0.9 and 0.999 respectively. Gradient clipping 5.0 was applied to avoid the gradient explosion problem. We trained all models with a learning rate $5e^{-4}$ and batch size 80. We decayed the learning rate with a factor of 0.5 between each training epoch. Translations were generated by a beam search algorithm that was based on log-likelihood scores normalized by sentence length. We used a beam size of 10 in all the experiments. We also applied dropout for English-German and English-French tasks on the output layer to avoid over-fitting, and the dropout rate was set to 0.2.

To train deep NMT models, we adopted the GNMT architecture~\cite{DBLP:journals/corr/WuSCLNMKCGMKSJL16}. We kept all the above settings, except the dimensionality of word embedding and hidden state which we set to be 512.

\subsection{Results on English-German Translation}

The translation results are shown in Table \ref{english_german_translation}. We also provide results of several existing systems that are trained with comparable experimental settings to ours. In particular, our single model yields a detokenized BLEU score of 21.99. In order to show that the proposed model can be orthogonal to previous methods that improve LSTM/GRU-based NMT, we integrate a single-layer context-aware (CA) encoder~\cite{8031316} into our system. The ATR+CA system further reaches 22.7 BLEU, outperforming the winner system~\cite{BUCK14.1097.L14-1074} by a substantial improvement of 2 BLEU points. Enhanced with the deep GNMT architecture, the GNMT+ATR system yields a gain of 0.89 BLEU points over the RNNSearch+ATR+CA and 1.6 BLEU points over the RNNSearch + ATR. Notice that different from our system which was trained on the parallel corpus alone, the winner system used a huge monolingual text to enhance its language model.  

\begin{table*}[t]
\begin{center}
{ \small
\begin{tabular}{l|l|c|c|c}
\multicolumn{1}{l|}{\bf System} &
\multicolumn{1}{l|}{\bf Architecture } &
\multicolumn{1}{c|}{\bf Vocab} &
\multicolumn{1}{c|}{\bf {\em tok} BLEU } &
\multicolumn{1}{c}{\bf {\em detok} BLEU}\\
\hline
\hline
\multicolumn{5}{c}{\it Existing end-to-end NMT systems} \\
\hline
\citet{jean-EtAl:2015:ACL-IJCNLP} & RNNSearch (GRU) + unk replace + large vocab & 500K & 34.11 & -\\
\citet{luong-EtAl:2015:ACL-IJCNLP} &  LSTM with 6 layers + PosUnk & 40K & 32.70 & - \\
\citet{DBLP:journals/corr/SutskeverVL14} & LSTM with 4 layers & 80K & 30.59 & - \\ 
\citet{DBLP:journals/corr/ShenCHHWSL15} & RNNSearch (GRU) + MRT + PosUnk & 30K & 34.23 & - \\
\citet{DBLP:journals/corr/ZhouCWLX16} & LSTM with 16 layers + F-F connections + 36M data & 80K & 37.70 & - \\
\citet{DBLP:journals/corr/WuSCLNMKCGMKSJL16} & LSTM with 8 layers + RL-refined WPM + 36M data & 32K & 38.95 & - \\
\citet{wang-EtAl:2017:Long1} & RNNSearch (GRU) with 4 layers + LAU & 30K & 35.10 & - \\
\citet{DBLP:journals/corr/GehringAGD16} & Deep Convolutional Encoder 20 layers with kernel width 5 & 30K & 35.70 & - \\
\citet{DBLP:journals/corr/VaswaniSPUJGKP17} & Transformer with 6 layers + 36M data + base model & 32K & 38.10 & - \\
\citet{DBLP:journals/corr/GehringAGYD17} & ConvS2S with 15 layers + 36M data & 40K & 40.46 & - \\
\citet{DBLP:journals/corr/VaswaniSPUJGKP17} & Transformer with 6 layers + 36M data + big model & 32K & 41.80 & - \\
\citet{DBLP:journals/corr/WuSCLNMKCGMKSJL16} & LSTM with 8 layers + RL WPM + 36M data + ensemble & 32K & 41.16 & - \\
\hline
\multicolumn{5}{c}{\it Our end-to-end NMT systems} \\
\hline
\multirow{6}{*}{\it this work} 
 									  & RNNSearch + GRU + BPE & 40K & 35.89 & 33.41 \\
									  & RNNSearch + LSTM + BPE & 40K & 36.95 & 34.15 \\
									  & RNNSearch + ATR + BPE & 40K & 36.89 & 34.00 \\
									  & RNNSearch + ATR + CA + BPE & 40K & 37.88 & 34.96 \\
									  & GNMT + ATR + BPE & 40K & 38.59 & 35.67 \\
									  & RNNSearch + ATR + CA + BPE + ensemble & 40K & 39.06 & 36.06 \\
\end{tabular}
}
\end{center}
\caption{\label{english_french_translation} Tokenized ({\em tok}) and detokenized (detok) case-sensitive BLEU scores on the WMT14 English-French translation task. ``{\em 12M data}'' indicates the same training data as ours, while ``{\em 36M data}'' is a significant larger dataset that contains the {\em 12M data}.}
\end{table*}
Compared with the existing LSTM-based~\cite{luong-pham-manning:2015:EMNLP} deep NMT system, our shallow/deep model achieves a gain of 2.41/3.26 tokenized BLEU points respectively. Under the same training condition, our ATR outperforms RAN by a margin of 0.34 tokenized BLEU points, and achieves competitive results against its GRU/LSTM counterpart. This suggests that although our ATR is much simpler than GRU, LSTM and RAN, it still possesses strong modeling capacity. 

In comparison to several advanced deep NMT models, such as the Google NMT (8 layers, 24.61 tokenized BLEU)~\cite{DBLP:journals/corr/WuSCLNMKCGMKSJL16} and the LAU-connected NMT (4 layers, 23.80 tokenized BLEU)~\cite{wang-EtAl:2017:Long1}, the performance of our shallow model (23.31) is competitive. Particularly, when replacing LSTM in the Google NMT with our ATR model, the GNMT+ATR system achieves a BLEU score of 24.16, merely 0.45 BLEU points lower. Notice that although all systems use the same training data of WMT14, the tokenization of these work might be different from ours. However, the overall results can indicate the competitive strength of our model. In addition, SRU~\cite{2017arXiv170902755L}, a recent proposed efficient recurrent unit, obtains a BLEU score of 20.70 with 10 layers, far more behind ATR's.

We further ensemble eight likelihood-trained models with different random initializations for the ATR+CA system. The variance in the  tokenized BLEU scores of these models is 0.07. As can be seen from Table \ref{english_german_translation}, the ensemble system achieves a tokenized and detokenized BLEU score of 24.97 and 24.33 respectively, obtaining a gain of 1.66 and 1.63 BLEU points over the single model. The final result of the ensemble system, to the best of our knowledge, is a very promising result that can be reached by single-layer NMT systems on WMT14 English-German translation.

\subsection{Results on English-French Translation}

Unlike the above translation task, the WMT14 English-French translation task provides a significant larger dataset. The full training data have approximately 36M sentence pairs, from which we only used 12M instances for experiments following previous work~\citep{jean-EtAl:2015:ACL-IJCNLP,DBLP:journals/corr/GehringAGD16,luong-EtAl:2015:ACL-IJCNLP,wang-EtAl:2017:Long1}. We show the results in Table \ref{english_french_translation}.

Our shallow model achieves a tokenized BLEU score of 36.89 and 37.88 when it is equipped with the CA encoder, outperforming almost all the listed systems, except the Google NMT~\cite{DBLP:journals/corr/WuSCLNMKCGMKSJL16}, the ConvS2S~\cite{DBLP:journals/corr/GehringAGYD17} and the Transformer~\cite{DBLP:journals/corr/VaswaniSPUJGKP17}. Enhanced with the deep GNMT architecture, the GNMT+ATR system reaches a BLEU score of 38.59, which beats the base model version of the Transformer by a margin of 0.49 BLEU points. When we use four ensemble models (the variance in the tokenized BLEU scores of these ensemble models is 0.16), the ATR+CA system obtains another gain of 0.47 BLEU points, reaching a tokenized BLEU score of 39.06, which is comparable with several state-of-the-art systems.

\begin{figure}[t]
	\centering
	\includegraphics[scale=0.45]{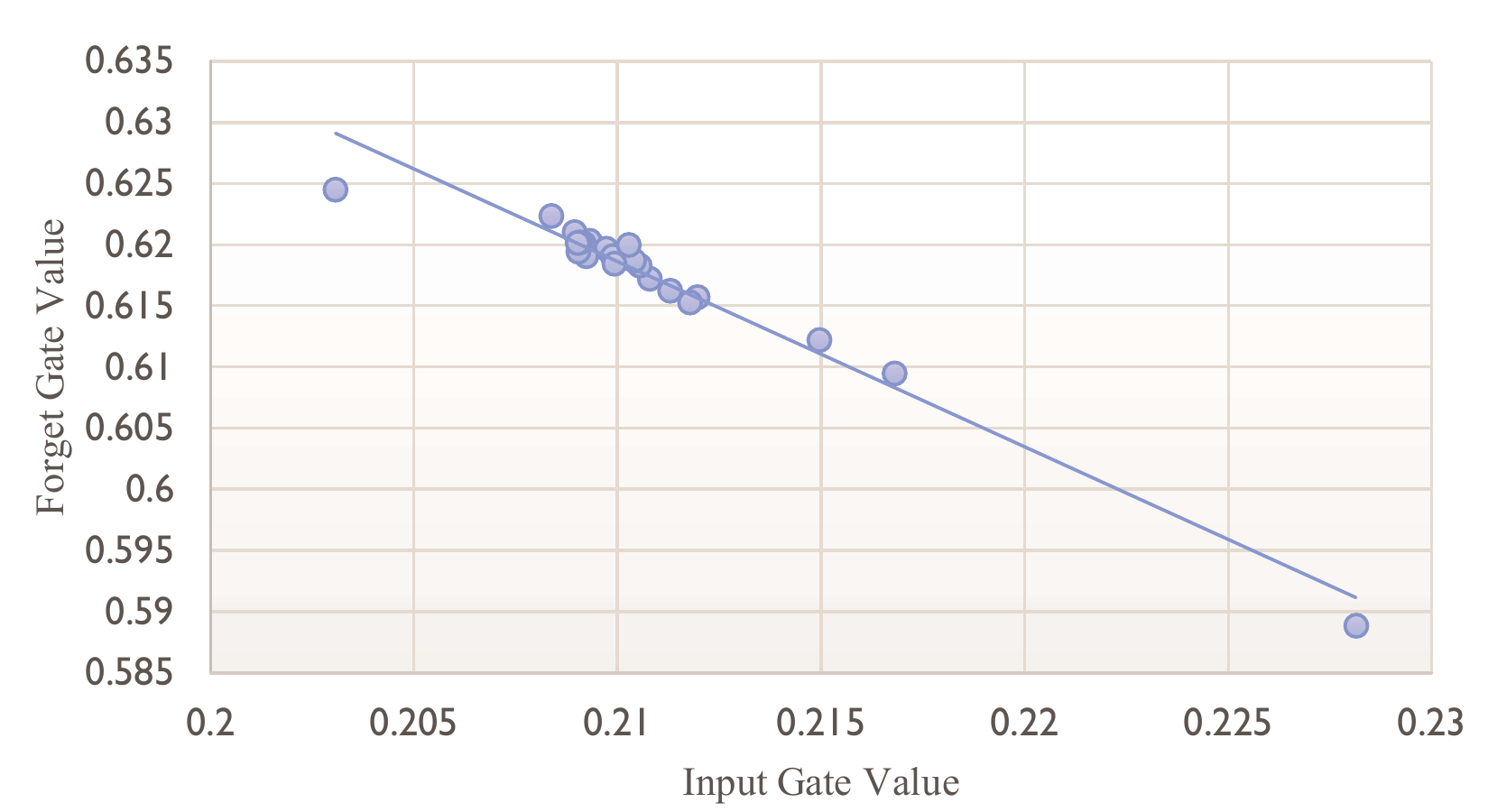}
  \caption{\label{visual_twin_mechanism} Visualization of correlation between the input and forget gate learned in Equation (\ref{decoder_higher_tsn}). For each gate, we record its mean value ($\frac{1}{d_h}\sum_i \mathbf{i}_{t,i}$/$\frac{1}{d_h}\sum_i \mathbf{f}_{t,i}$) with its position on newstest2014, and then average all gate values for each position over all test sentences. This figure illustrates the position-wise input and forget gate values.}
\end{figure}

\begin{figure*}[t]
  \centering
  \includegraphics[scale=0.43]{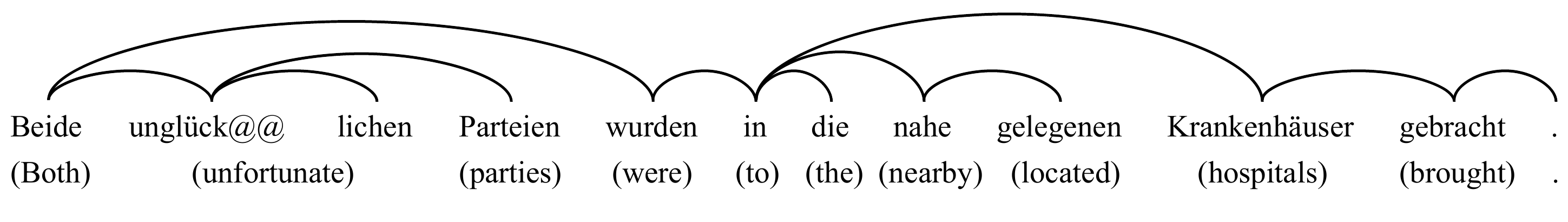}
  \caption{\label{dep_linesss} Visualization of dependencies on a target (German) sentence (selected from newstest2014). ``@@'' indicates a separator that splits one token into two pieces of sub-words.}
\end{figure*}
\section{Analysis}\label{exp_analysis}

\subsection{Analysis on Twin-Gated Mechanism}\label{exp_analysis_twin}

We provide an illustration of the actual relation between the learned input and forget gate in Figure \ref{visual_twin_mechanism}. Clearly, these two gates show strong negative correlation. When the input gate opens with high values, the forget gate prefer to be close. Quantitatively, on the whole test set, the Pearson's r of the input and forget gate is -0.9819, indicating a high correlation. 

\subsection{Analysis on Speed and Model Parameters}\label{exp_analysis_speed}

As mentioned in Section \ref{com_analysis}, ATR has much fewer model parameters and matrix transformations. We provide more details in this section by comparing against the following two NMT systems:
\begin{itemize}
\item
{\em DeepRNNSearch (GRU)}: a deep GRU-equipped RNNSearch model~\cite{DBLP:journals/corr/WuSCLNMKCGMKSJL16} with 5 layers. We set the dimension of word embedding and hidden state to 620 and 1000 respectively.
\item
{\em Transformer}: a purely attentional translator~\cite{DBLP:journals/corr/VaswaniSPUJGKP17}. We set the dimension of word embedding and filter size to 512 and 2048 respectively. The model was trained with a minibatch size of 256.
\end{itemize}
We also compare with the GRU and LSTM-based RNNSearch. Without specific mention, all other experimental settings for all these models are the same as for our model. We implement all these models using the {\em Theano} library, and test the speed on one GeForce GTX TITAN X GPU card. We show the results on Table \ref{model_analysis}.

\begin{table}[t]
\begin{center}
{\small
\begin{tabular}{l|l|l|l}
{\bf Model} & {\bf \#PMs} & {\bf Train} & {\bf Test} \\
\hline
\hline
{\it RNNSearch+GRU} & 83.5M & 1996 & 168 (0.133) \\
{\it RNNSearch+LSTM} & 93.3M & 1919 & 167 (0.139) \\
{\it RNNSearch+RAN} & 79.5M & 2192 & 170 (0.129) \\
{\it DeepRNNSearch} & 143.0M & 894 &  70 (0.318) \\
{\it Transformer} & 70.2M & 4961 & 44 (0.485) \\
{\it RNNSearch+ATR} & 67.8M & 2518 & 177 (0.123) \\
{\it RNNSearch+ATR+CA} & 63.1M & 3993 & 186 (0.118) \\
\end{tabular}
}
\end{center}
\caption{\label{model_analysis} Comparison on the training and decoding speed and the number of model parameters of different NMT models on WMT14 English-German translation task with beam size 1. {\bf \#PMs}: the number of model parameters. {\bf Train/Test}: the number of words in one second processed during training/testing. The number in bracket indicates the average decoding time per source sentence (in seconds).}
\end{table}
We observe that the Transformer achieves the best training speed, processing 4961 words per second. This is reasonable since the  Transformer can be trained in full parallelization. On the contrary, DeepRNNSearch is the slowest system. As RNN performs sequentially, stacking more layers of RNNs inevitably reduces the training efficiency. However, this situation becomes the reverse when it comes to the decoding procedure. The Transformer merely generates 44 words per second while DeepRNNSearch reaches 70. This is because during decoding, all these beam search-based systems must generate translation one word after another. Therefore the parallelization advantage of the Transformer disappears. In comparison to DeepRNNSearch, the Transformer spends extra time on performing self-attention over all previous hidden states.

Our model with the CA structure, using only 63.1M parameters, processes 3993 words per second during training and generates 186 words per second during decoding, which yields substantial speed improvements over the GRU- and LSTM-equipped RNNSearch. This is due to the light matrix computation in recurrent units of ATR. Notice that the speed increase of ATR over GRU and LSTM does not reach 3x. This is because at each decoding step, there are mainly two types of computation: recurrent unit and softmax layer. The latter consumes the most calculation, which, however, is the same for different models (LSTM/GRU/ATR).

\subsection{Analysis on Dependency Modeling}\label{exp_analysis_dep}

As shown in Section \ref{dependency_analysis}, a hidden state in our ATR can be formulated as a weighted sum of the previous inputs. In this section, we quantitatively analyze the weights $\mathbf{g}_k$ in Equation (\ref{dep_gate_value}) induced from Equation (\ref{decoder_higher_tsn}). Inspired by~\newcite{DBLP:journals/corr/LeeLZ17}, we visualize the captured dependencies of an example in Figure \ref{dep_linesss} where we connect each word to the corresponding previous word with the highest weight $\mathbf{g}_k$.

Obviously, our model can discover strong local dependencies. For example, the token ``{\em unglück@@}'' and ``{\em lichen}'' should be a single word. Our model successfully associates ``{\em unglück@@}'' closely to the generation of ``{\em lichen}'' during decoding. In addition, our model can also detect non-consecutive long-distance dependencies. Particularly, the prediction of ``{\em Parteien}'' relies heavily on the token ``{\em unglücklichen}'', which actually entails an {\em amod} linguistic dependency relationship. These captured dependencies make our model more interpretable than LSTM/GRU.

\section{Conclusion and Future Work}

This paper has presented a twin-gated recurrent network (ATR) to simplify neural machine translation. There are only two weight matrices and matrix transformations in recurrent units of ATR, making it efficient in physical memory usage and running speed. To avoid the gradient vanishing problem, ATR introduces a twin-gated mechanism to generate an input gate and forget gate through linear addition and subtraction operation respectively, without introducing any additional parameters. The simplifications allow ATR to produce interpretable results.

Experiments on English-German and English-French translation tasks demonstrate the effectiveness of our model. They also show that ATR can be orthogonal to and applied with methods that improve LSTM/GRU-based NMT, indicated by the promising performance of the ATR+CA system. Further analyses reveal that ATR can be trained more efficiently than GRU. It is also able to transparently model long-distance dependencies.  

We also adapt our ATR to other natural language processing tasks. Experiments show encouraging performance of our model on Chinese-English translation, natural language inference and Chinese word segmentation, demonstrating its generality and applicability on various NLP tasks.

In the future, we will continue to examine the effectiveness of ATR on different neural models for NMT, such as the hierarchical NMT model~\cite{su2018hierarchy} as well as the generative NMT model~\cite{su2018variational}. We are also interested in adapting our ATR to summarization, semantic parsing etc.

\section*{Acknowledgments}

The authors were supported by National Natural Science Foundation of China (Grants No. 61672440, 61622209 and 61861130364), the Fundamental Research Funds for the Central Universities (Grant No. ZK1024), and Scientific Research Project of National Language Committee of China (Grant No. YB135-49). Biao Zhang greatly acknowledges the support of the Baidu Scholarship. We also thank the reviewers for their insightful comments.

\bibliography{emnlp2018}
\bibliographystyle{acl_natbib_nourl}

% \begin{comment}
\clearpage
\appendix

\section{Appendix}

\begin{table*}[t]
\begin{center}
{ \small
\begin{tabular}{l|c|cllllll}
\multicolumn{1}{l|}{\bf System} &
\multicolumn{1}{c|}{\bf MT05 } &
\multicolumn{1}{l}{\bf MT02 } &
\multicolumn{1}{l}{\bf MT03 } &
\multicolumn{1}{l}{\bf MT04 } &
\multicolumn{1}{l}{\bf MT06 } &
\multicolumn{1}{l}{\bf MT08 } \\
\hline
\hline
\multicolumn{7}{c}{\em Existing Systems} \\
\hline
{\it Coverage~\citep{2017arXiv170500861W}} & 34.91 & - & 34.49 & 38.34 & 34.25 & - \\
{\it MemDec~\citep{2017arXiv170500861W}} & 35.91 & - & 36.16 & 39.81 & 35.98 & - \\
{\it DeepLAU~\citep{2017arXiv170500861W}} & 38.07 & - & 39.35 & 41.15 & 37.29 & -\\
{\it Distortion~\citep{zhang-EtAl:2017:Long3}} & 36.71 & - & 38.33 & 40.11 & 35.29 & - \\
{\it CAEncoder~\citep{8031316}} & 36.44 & 40.12 & 37.63 & 39.83 & 35.44 & 27.34\\
{\it FPNMT~\citep{DBLP:journals/corr/abs-1711-09502}} & 36.75 & 39.65 & 37.90 & 40.37 & 34.55 & - \\
{\it ASDBNMT~\citep{DBLP:journals/corr/abs-1801-05122}} & 38.84 & - & 40.02 & 42.32 & 38.38 & - \\ 
\hline
\multicolumn{7}{c}{\em Our end-to-end NMT systems} \\
\hline
{\it this work} & {\bf 39.71} & {\bf 42.95} & {\bf 41.71} & {\bf 43.71} & {\bf 39.61} & {\bf 31.14}
\end{tabular}
}
\end{center}
\caption{\label{c2e_comparison} Case-insensitive BLEU scores of advanced systems on the Chinese-English translation tasks. ``$-$'' indicates that no result is provided in the original paper.}
\end{table*}
\subsection{Neural Machine Translation with ATR}\label{nmt_atr}

We replace LSTM/GRU with our proposed ATR to build NMT models under the attention-based encoder-decoder framework~\citep{DBLP:journals/corr/BahdanauCB14}. The encoder that reads a source sentence is a bidirectional recurrent network. Formally, given a source sentence $\mathbf{x}=\{\mathbf{x}_1, \mathbf{x}_2, \ldots, \mathbf{x}_n\}$, the encoder is formulated as follows:
\begin{equation}
\begin{split}
\overrightarrow{\mathbf{h}}_{i} = \text{ATR}(\overrightarrow{\mathbf{h}}_{i-1}, \mathbf{x}_i), 
\overleftarrow{\mathbf{h}}_{i} = \text{ATR}(\overleftarrow{\mathbf{h}}_{i+1}, \mathbf{x}_i)
\end{split}
\end{equation}
where $\text{ATR}(\cdot)$ is defined by Equation (\ref{eq_eru}\&\ref{twin_h}). The forward $\overrightarrow{\mathbf{h}}_i$ and backward $\overleftarrow{\mathbf{h}}_i$ hidden states are concatenated together to represent the $i$-th word: $\mathbf{h}_i = [\overrightarrow{\mathbf{h}}_i; \overleftarrow{\mathbf{h}}_i]$. %, which is successively fed into a recurrent decoder. 

The decoder is a conditional language model that predicts the $j$-th target word via a multilayer perception:
\begin{equation}
p(y_j|\mathbf{x}, \mathbf{y}_{<j}) = \text{softmax}(g(\mathbf{y}_{j-1}, \tanh(\mathbf{s}_j), \mathbf{c}_j))
\end{equation}
where $\mathbf{y}_{<j}$ is a partial translation. $\mathbf{c}_j$ is the translation-sensitive semantic vector computed via the attention mechanism~\citep{DBLP:journals/corr/BahdanauCB14} based on the source states $\{\tanh(\mathbf{h}_{i})\}_{i=1}^n$ and internal target state $\widetilde{\mathbf{s}}_{j}$, and $\mathbf{s}_j$ is the $j$-th target-side hidden state calculated through a two-level hierarchy:
\begin{align}
\widetilde{\mathbf{s}}_j = \text{ATR}(\mathbf{s}_{j-1}, \mathbf{y}_{j-1}), \mathbf{s}_j = \text{ATR}(\widetilde{\mathbf{s}}_j, \mathbf{c}_j) \label{decoder_higher_tsn}
\end{align}

\subsection{Additional Experiments}~\label{other_exp}
\subsubsection{Experiments on Chinese-English Translation}

Our training data consists of 1.25M sentence pairs including 27.9M Chinese words and 34.5M English words respectively.\footnote{This corpora contain LDC2002E18, LDC2003E07, LDC2003E14, Hansards portion of LDC2004T07, LDC2004T08 and LDC2005T06.} We used the NIST 2005 dataset as our dev set, and the NIST 2002, 2003, 2004, 2006 and 2008 datasets as our test sets. Unlike WMT14 translation tasks, we used word-based vocabulary for Chinese-English, preserving top-30K most frequent source and target words in the vocabulary. Case-insensitive BLEU-4 metric was used to evaluate the translation quality.

\noindent{\bf Translation Results}

We compare our model against several advanced models on the same dataset, including:
\begin{itemize}
\item
{\it Coverage~\citep{2017arXiv170500861W}}: an attention-based NMT system enhanced with a coverage mechanism to handle the over-translation and under-translation problem.
\item
{\it MemDec~\citep{2017arXiv170500861W}}: an attention-based NMT system that replaces the vanilla decoder with a memory-enhanced decoder to better capture important information for translation.
\item
{\it DeepLAU~\citep{2017arXiv170500861W}}: a deep attention-based NMT system integrated with linear associative units that deals better with gradient propagation.
\item
{\it Distortion~\citep{zhang-EtAl:2017:Long3}}: an attention-based NMT system that incorporates word reordering knowledge to encourage more accurate attention.
\item
{\it CAEncoder~\citep{8031316}}: the same as our model but uses GRU unit.
\item
{\it FPNMT~\citep{DBLP:journals/corr/abs-1711-09502}}: an attention-based NMT system that leverages past and future information to improve the attention model and the decoder states, also using addition and subtraction operations.
\item
{\it ASDBNMT~\citep{DBLP:journals/corr/abs-1801-05122}}: an attention-based NMT system that is equipped with a backward decoder to explore bidirectional decoding.
\end{itemize}
Table \ref{c2e_comparison} summarizes the results. Although our model does not involve any sub-networks for modeling the coverage, distortion,  memory and future context, our model clearly outperforms all these advanced models, achieving an average BLEU score of 39.82 on all test sets. This strongly suggests that 1) shallow models are also capable of generating extremely high-quality translations, and 2) our ATR model indeed has the ability in capturing translation correspondence in spite of its simplicity.

\subsubsection{Experiments on Natural Language Inference}

Given two sentences, namely a premise and a hypothesis, this task aims at recognizing whether the premise can entail the hypothesis. We used the Stanford Natural Language Inference Corpus (SNLI)~\cite{snli:emnlp2015} for this experiment, which involves a collection of 570k human-written English sentence pairs manually labeled for balanced classification with the labels {\em entailment}, {\em contradiction}, and {\em neutral}. We formulated this problem as a three-way classification task.

We employed the attentional architecture~\cite{rocktaschel2016reasoning} as our basic model, and replaced its recurrent unit with our ATR model. We fixed word embedding initialized with the pre-trained 300-D Glove vector~\cite{pennington2014glove}. The hidden size of ATR was also set to 300. We optimized model parameters using the Adam method~\cite{DBLP:journals/corr/KingmaB14} with hyperparameters $\beta_1=0.9$ and $\beta_2=0.999$. The learning rate was fixed at 0.0005. The mini-batch size was set to 128. Dropout was applied on both word embedding layer and pre-classification layer to avoid overfitting, with a rate of 0.15. The maximum training epoch was set to 20. 

\noindent {\bf Classification Results}

\begin{table*}[t]
\begin{center}
{\small
\begin{tabular}{l|cccc}
{\bf Model} & {\bf $d$} & {\bf \#Params} & {\bf Train} & {\bf Test} \\
\hline
\hline

LSTM encoders~\cite{bowman-EtAl:2016:P16-1} & 300 & 3.0m & 83.9 & 80.6 \\
GRU encoders w/ pretraining~\cite{DBLP:journals/corr/VendrovKFU15} & 1024 & 15m & 98.8 & 81.4 \\
BiLSTM encoders with intra-attention~\cite{DBLP:journals/corr/LiuSLW16} & 600 & 2.8m & 84.5 & 84.2 \\
LSTMs w/ word-by-word attention~\cite{rocktaschel2016reasoning} & 100 & 250k & 85.3 & 83.5 \\
mLSTM word-by-word attention model~\cite{wang-jiang:2016:N16-1} & 300 & 1.9m & 92.0 & 86.1 \\
LSTMN with deep attention fusion~\cite{cheng-dong-lapata:2016:EMNLP2016} & 450 & 3.4m & 88.5 & 86.3 \\
BiMPM~\cite{DBLP:journals/corr/WangHF17} & 100 & 1.6m & 90.9 & 87.5 \\
\hline
\hline

this work with GRU & 300 & 3.2m & 91.0 & 84.6 \\
this work with ATR & 300 & 1.5m & 90.9 & 85.6 \\

\end{tabular}
}
\end{center}
\caption{\label{snli_result} Classification results on the SNLI task. For comparison, we provide the model dimension ($d$), the parameter amount (\# Params), the training accuracy (Train) and the test accuracy (Test). $m$: million. We also provide results of several existing RNN models from the SNLI official website.}
\end{table*}

\begin{table*}[t]
\begin{center}
{\small
\begin{tabular}{l|ccc|ccc}
\multirow{2}{*}{\bf Model} & \multicolumn{3}{|c|}{\bf MSRA} & \multicolumn{3}{c}{\bf CTB6} \\
\cline{2-7}
 & P & R & F & P & R & F \\
\hline
\citep{zheng2013deep} & 92.9 & 93.6 & 93.3 & 94.0 & 93.1 & 93.6 \\
\citep{pei-ge-chang:2014:P14-1} & 94.6 & 94.2 & 94.4 & 94.4 & 93.4 & 93.9 \\
\citep{chen2015long} & 96.7 & 96.2 & 96.4 & 95.0 & 94.8 & 94.9 \\
\hline
this work + LSTM & 95.5 & 94.9 & 95.2 & 93.3 & 93.1 & 93.2 \\
this work + GRU & 95.2 & 95.1 & 95.1 & 93.3 & 93.0 & 93.2 \\
this work + ATR & 95.3 & 95.1 & 95.2 & 94.0 & 93.9 & 94.0
\end{tabular}
}
\end{center}
\caption{\label{cws_result} Model performance on MSRA and CTB6 datasets. We report precision (P), recall (R) and F1-score (F).}
\end{table*}
Table \ref{snli_result} shows the results. The GRU equipped model in our implementation achieves a test accuracy of 84.6\% with about 3.2m trainable model parameters, outperforming the LSTM-enhanced counterpart~\cite{rocktaschel2016reasoning} by a margin of 1.1\%. By contrast, the same architecture with ATR model yields a test accuracy of 85.6\%, with merely 1.5m model parameters. In other words, using fewer parameters, our ATR model gains a significant improvement of 1.0\%, reaching a comparable performance against some deep architectures~\cite{cheng-dong-lapata:2016:EMNLP2016}. % This reveals that ATR is capable of manipulating its limited parameters to model the underlying task.

\subsubsection{Experiments on Chinese Word Segmentation}

Chinese word segmentation (CWS) is a fundamental preprocessing step for Chinese-related NLP tasks. Unlike other languages, Chinese sentences are recorded without explicit delimiters. Therefore, before performing in-depth modeling, researchers need to segment the whole sentence into a sequence of tokens, which is exactly the goal of CWS. 

Following previous work~\cite{chen2015long}, we formulate CWS as a sequence labeling task. Each character in a sentence is assigned with a unique label from the set {\em \{B, M, E, S\}}, where {\em \{B, M, E\}} indicate {\em Begin, Middle, End} of a multi-character word respectively, and {\em S} denotes a {\em Single} character word. Given a sequence of characters, we first embed them individually through a character embedding layer, followed by a bidirectional RNN layer to generate context-sensitive representation for each character. The output representations are then passed through a CRF inference layer to capture dependencies among character labels. The whole model is optimized using a max-margin objective towards minimizing the differences between predicted sequences and gold label sequences.  

We used the MSRA and CTB6 datasets to evaluate our model. The former is provided by the second International Chinese Word Segmentation Bakeoff~\cite{Sproat:2003:FIC:1119250.1119269}, and the latter is from Chinese TreeBank6.0 (LDC2007T36)~\cite{Xue:2005:PCT:1064781.1064785}. For MSRA dataset, we split the first 90\% sentences of the training data as the training set and the rest as the development set. For CTB6 dataset, we divided the training, development and test sets in the same way as in~\cite{chen2015long}. Precision, recall, F1-score and out-of-vocabulary (OOV) word recall calculated by the standard back-off scoring program were used for evaluation.

We set the dimensionality of both character embedding and RNN hidden state to be 300. Model parameters were tuned by Adam algorithm~\cite{DBLP:journals/corr/KingmaB14} with default hyperparameters ($\beta_1=0.9, \beta_2=0.999$) and mini-batch size 128. Gradient was clipped when its norm exceeds 1.0 to avoid gradient explosion. We applied dropout on both character embedding layer and pre-CRF layer with a rate of 0.2. The discount parameter in max-margin objective was set to 0.2. The maximum training epoch was set to 50. Learning rate was initially set to 0.0005, and halved after each epoch.

\noindent{\bf Model Performance}

Table \ref{cws_result} shows the overall performance. We observe that our ATR model performs as efficient as both GRU and LSTM on this  task. ATR yields a F1-score of 95.2\% and 94.0\% on MSRA and CTB6 dataset respectively, almost the same as that of GRU (95.1\%/93.2\%) and LSTM (95.2\%/93.2\%). Particularly, ATR achieves better results on CTB6, with a gain of 0.8\% F1 points over GRU and LSTM. This further demonstrates the effectiveness of the proposed ATR model. 
% \end{comment}

\end{document}